\documentclass[]{article}
\usepackage{style}

\usepackage[utf8]{inputenc} % allow utf-8 input
\usepackage[T1]{fontenc}    % use 8-bit T1 fonts
\usepackage{hyperref}       % hyperlinks
\usepackage{url}            % simple URL typesetting
\usepackage{booktabs}       % professional-quality tables
\usepackage{amsfonts}       % blackboard math symbols
\usepackage{nicefrac}       % compact symbols for 1/2, etc.
\usepackage{microtype}      % microtypography
\usepackage{lipsum}
\usepackage{fancyhdr}       % header
\usepackage{graphicx}       % graphics
\usepackage{multicol}
\usepackage{longtable}
\usepackage{tabularx}
\usepackage{float}
\usepackage{caption}
\usepackage{setspace} 

\usepackage[
backend=biber,
style=ieee,
citestyle=numeric-comp
]{biblatex}

\graphicspath{{media/}} 

\title{Context-Aware Prediction of User Engagement on Online Social Platforms}

\author{
Heinrich Peters            \\
Columbia University\\
\texttt{\ hp2500@columbia.edu}\\
\And
Yozen Liu\\
Snap Inc. \\
\texttt{\ yliu2@snap.com}\\
\And
Francesco Barbieri\\
Snap Inc.\\
\texttt{\ fbarbieri@snap.com}\\
\AND
Raiyan Abdul Baten\\
University of South Florida\\
\texttt{\ rbaten@usf.edu}\\
 \And
Sandra C. Matz\\
Columbia University\\
\texttt{\ sm4409@columbia.edu}\\
\And
Maarten W. Bos\\
Snap Inc.\\
\texttt{\ maarten@snap.com}\\
}

\begin{document}

\maketitle
% \doublespacing
\begin{abstract}
The success of online social platforms hinges on their ability to predict and understand user behavior at scale. Here, we present data suggesting that context-aware modeling approaches may offer a holistic yet lightweight and potentially privacy-preserving representation of user engagement on online social platforms. Leveraging deep LSTM neural networks to analyze more than 100 million Snapchat sessions from almost 80.000 users, we demonstrate that patterns of active and passive use are predictable from past behavior ($R^2$=0.345) and that the integration of context features substantially improves predictive performance compared to the behavioral baseline model ($R^2$=0.522). Features related to smartphone connectivity status, location, temporal context, and weather were found to capture non-redundant variance in user engagement relative to features derived from histories of in-app behaviors. Further, we show that a large proportion of variance can be accounted for with minimal behavioral histories if momentary context is considered ($R^2$=0.442). These results indicate the potential of context-aware approaches for making models more efficient and privacy-preserving by reducing the need for long data histories. Finally, we employ model explainability techniques to glean preliminary insights into the underlying behavioral mechanisms. Our findings are consistent with the notion of context-contingent, habit-driven patterns of active and passive use, underscoring the value of contextualized representations of user behavior for predicting user engagement on social platforms.
\end{abstract}

\vspace{10mm}

% keywords can be removed

\newpage
%\begin{multicols}{2}
\section{Introduction}
Few technologies have shaped the world as drastically as the advent of the internet and the rise of online social networks. As people's lives are increasingly mediated through online platforms competing for users’ attention and striving to maximize user engagement, the ability to understand and predict user behavior has become vitally important. This capacity forms the basis of powerful personalization technologies that adapt to individual users’ needs in order to provide a better user experience and ultimately generate revenue. 
The impact of such technologies can hardly be overestimated as they are deeply embedded in the design - and will continue to be at the core of the business model - of online platforms that billions of people use on a daily basis. 
The capability to predict user behavior at increasing levels of granularity, however, comes at a cost. Not only does it require increasingly sophisticated infrastructure and software, it can also compromise users’ privacy \cite{casadesus-masanell_strategies_2020, chellappa_personalization_2005, cloarec_personalizationprivacy_2020, gal-or_role_2018} as the prevailing approach to modeling user behavior involves the creation of user-profiles and long histories of past behavior to predict future behavior \cite{eke_survey_2019}, often across platforms \cite{veiga_cross-platform_2016}. While the design of predictive technologies tends to be agnostic to theoretical insights from the behavioral sciences, their application is closely aligned with the catchphrase that past behavior is the best predictor of future behavior \cite{aarts_predicting_1998, ouellette_habit_1998}. However, the availability of fine-grained behavioral user data also opens the door for modeling approaches aimed at deeper psychological constructs, such as personality \cite{kosinski_private_2013, kulkarni_latent_2018, matz_psychological_2017, youyou_computer-based_2015, peters_large_2023, peters_large_2024} or habits \cite{chowdhury_ceam_2021, liu_characterizing_2019, peters_social_2024}. 

In this paper, we propose that a context-aware approach to user modeling can increase the performance of predictive models while deepening our understanding of online social behavior. That is, in building contextualized, holistic representations of user engagement, predictive models can be made more efficient, more privacy-preserving, and more aligned with behavioral theory than current approaches. 
Specifically, we employ LSTM neural networks \cite{schmidhuber_deep_2015, hochreiter_long_1997}, which excel at representing the recurrent behavioral patterns characteristic of media and technology habits \cite{liu_characterizing_2019, schnauber-stockmann_process_2019, tokunaga_media_2020, chowdhury_ceam_2021, peters_social_2024}, to explore how context-aware models incorporating smartphone connectivity status, location, temporal context, weather, and socio-demographic context can aid the prediction of active and passive user behaviors in online social platforms.

% \subsection{Related Research}
Recognizing that user engagement is a multi-faceted phenomenon, past research has often differentiated between active and passive user behaviors. Active use includes behaviors that revolve around social interactions and the creation of content, while passive use includes behaviors that revolve around content consumption \cite{escobar-viera_passive_2018, hemmings-jarrett_evaluation_2017, khan_social_2017, pagani_influence_2011, trifiro_social_2019}. The distinction between active and passive use is of both theoretical and practical relevance. While prior work has examined the impact of active versus passive use on psychological outcomes such as well-being \cite{escobar-viera_passive_2018, verduyn_social_2022}, little is known about habitual, context-contingent patterns of active and passive use on social platforms. At the same time, the ability to predict active and passive use is important for businesses because the two modes of user engagement distinctively impact user experience and revenue. For example, it is important for online social platforms to host a variety of user-generated content in order to keep users socially engaged \cite{pagani_influence_2011}, but passive use has become increasingly relevant because ad revenue is closely tied to content consumption on all major platforms as content is interspersed with ads. Relatedly, the recent rise of short-form video content has strongly affected how people interact with online social platforms, and the context-contingent and potentially habitual nature of these novel forms of user behavior is currently not well understood.

The field of psychology has produced a rich body of work regarding the effects of context on human behavior. This includes research on ecological psychology \cite{barker_explorations_1965, heft_ecological_2016, lobo_history_2018}, person-situation interactions \cite{rauthmann_person-situation_2020, funder_persons_2009, fleeson_end_2008}, the conceptualization and measurement of situational cues \cite{rauthmann_conceptualizing_2021, rauthmann_situational_2014, schoedel_snapshots_2023, funder_taking_2016}, as well as media and technology habits \cite{bayer_building_2022, verplanken_technology_2018, anderson_habits_2021}. With regard to the prediction of user engagement, the literature on media and technology habits provides the most suitable theoretical framework connecting context and online behavior. Media and technology habits are learned routines that are contingent on contextual cues and emerge after repeated media use \cite{schnauber-stockmann_process_2019, tokunaga_media_2020, bayer_building_2022}. Habit formation is initially driven by reward learning, where the pleasure associated with content consumption, or the experience of social interaction and recognition, act as rewards. Once a habit is formed, the initial motivations become less important, and the learned response occurs automatically when triggered by associated contextual cues \cite{anderson_habits_2021}. For example, a user might habitually scroll through social media when riding the bus to work or open an instant messaging app when they hear a notification sound. Generally, contextual cues can be conceptualized as specific context features that precede habitualized behaviors \cite{anderson_habits_2021, gardner_habit_2019}. Relevant cues include “technical cues” that are features of a medium or technology itself (e.g., notifications, buzzes, sounds) \cite{verplanken_technology_2018}, but also more subtle situational triggers, such as the time of the day, properties of the environment, events and activities, or internal states \cite{larose_problem_2010, naab_habitual_2016, verplanken_interventions_2006, wood_new_2007, bayer_connection_2016}. 

% TODO ADD EXPLANATION ABOUT CONTEXT

Previous empirical research indicates that more than half of media consumption can be classified as habitual \cite{wood_habits_2002} and that repeated social media use is strongly habitual, with frequent users reporting higher degrees of automaticity \cite{anderson_habits_2021}. Habitual use has also been found to be the most important predictor of media use for social sharing \cite{choi_understanding_2021} and news consumption \cite{diddi_getting_2006}. 
In line with these findings, recent research has shown that user activity on social media apps follows predictable patterns. For example, user engagement on Snapchat follows habitual patterns over time, enabling new neural network architectures specifically designed to represent cyclical patterns \cite{chowdhury_ceam_2021}. Similarly, fine-grained in-app action sequences follow predictable patterns that can be characterized as habitual \cite{liu_characterizing_2019}.
However, with the exception of time, the role of context has largely been overlooked in studies linking habits and user engagement.

While there is a lack of previous research investigating how active and passive use relate to contextualized habits, the computer science literature on context-aware modeling offers valuable insights linking user activity and context. The notion of context-awareness - the “ability of a mobile user’s applications to discover and react to changes in the environment” \cite{schilit_disseminating_1994} - shares its emphasis on context with the habits literature but focuses more on applications than on behavioral theory \cite[e.g.,][]{yurur_context-awareness_2016, sarker_context-aware_2021}. Recently, there has been a strong movement towards utilizing machine learning in context-aware technology to model user behavior and tailor applications to individual users' needs \cite[e.g.,][]{nascimento_context-aware_2018, raza_progress_2019, mijnsbrugge_context-aware_2023, sarker_context-aware_2021, salido_ortega_towards_2020, miranda_survey_2022}. For instance, novel methods have been developed to predict users' activities \cite{liao_predicting_2018, hasan_context_2015}, locations \cite{liao_predicting_2018, baten_predicting_2023}, and mental states \cite{lam_context-aware_2019, muller_depression_2021, onim_review_2023, salido_ortega_towards_2020}, as well as interactions with mobile applications \cite{chowdhury_ceam_2021, fan_personalized_2019, sarker_context-aware_2021, xia_deepapp_2020, huang_predicting_2012}. Similarly, context-aware recommender systems have received attention in past research \cite{raza_progress_2019, chen_context-aware_2005}. A wide array of context features has proven useful for the purpose of context-aware modeling, including spatial and temporal context \cite{fan_personalized_2019, xia_deepapp_2020}, and device context \cite{sarker_context-aware_2021}, but also more indirect factors such as weather \cite{nawshin_modeling_2020}. Despite these advancements, it has been pointed out that the collection, processing, and modeling of various context features remains a challenge, especially in mobile settings \cite{sarker_mobile_2021}. 

Notably, past work has not always been consistent with regard to its definition of context. For example, the literature on context-aware computing tends to operate on a broader notion of context compared to the behavioral science literature. The behavioral science literature usually defines context as the sum of environmental conditions or situational circumstances under which behavior occurs, explicitly juxtaposing context-characteristics and person-characteristics or inner states as distinct (and sometimes competing) determinants of behavior \cite[e.g.,][]{sansone_sage_2004}. This definition is aligned with the behavioral sciences’ focus on individual behavior as the main object of study. The computer science literature, on the other hand, tends to include person characteristics such as preferences, psychological traits, and even biometric processes in their definitions of context \cite[e.g.,][]{gajjar_context-aware_2017}. This perspective is aligned with a technology-centric view in which user characteristics are part of the context in which a technological system is studied. Since the present research aims to contribute not only to the technical literature on context-aware computing but also to the behavioral science literature, it employs the former definition of context, highlighting the dichotomy between environmental factors and person characteristics. Similarly, while past research has sometimes characterized preceding behavior as context \cite{wood_habit_2017, peters_social_2024, buyalskaya_what_2023}, the present paper operates on a more narrow definition of context as the sum of environmental factors, focusing on geographic areas with certain socio-demographic properties, weather, time, and locations or places. This approach captures a comprehensive view of spatial context, including factors that are relatively constant and mostly outside users' control (socio-demographic context and weather), along with those that more closely represent users' decisions and mobility habits (location visits and connectivity status).

% \subsection{Current Research}
The present research explores how context-aware models representing habitual patterns of behavior can aid the prediction of active and passive user behaviors. We analyze data from Snapchat, a major instant messaging and online social platform with close to 400 million daily active users \cite{snap_inc_snap_2023}. In the US, the majority (65\%) of 18-29 year-olds use Snapchat, most of them at least daily \cite{auxier_social_2021}. The app provides ample opportunities for active and passive user behaviors. A core functionality of the Snapchat app is messaging, allowing users to exchange written notes (chats), photos and videos (Snaps), or other multimedia content found on the app. Messaging is inherently social and therefore constitutes a prime example of active use. Other examples of active use include the Lens feature, which allows users to creatively edit photos and videos before sharing them, and the Stories feature allowing to post multimedia status updates that can be viewed by some or all of a user’s friends. Examples of passive use include the Discover Feed and the Spotlight Feed, which provide a curated collection of multimedia content that users can engage with. The large user base, the high frequency of interaction, and the design of the platform allowing for a diverse set of active and passive user behaviors make Snapchat an ideal setting to study the prediction of user engagement.

Based on data from over 79,000 Snapchat users, we investigate the extent to which the integration of contextual information, including smartphone connectivity status, location, temporal context, weather, and socio-demographic context, can lead to more accurate, efficient, and privacy-preserving models of user engagement. Using deep LSTM neural networks \cite{schmidhuber_deep_2015, hochreiter_long_1997}, we first show that - consistent with the idea of habit-driven user engagement - active and passive use follow predictable patterns over time ($R^2$=0.345). Second, highlighting the context-contingent nature of media and technology habits, we demonstrate that the model can be improved by adding different types of context features ($R^2$=0.522). Third, we show that very short sequences can lead to satisfactory predictive performance if momentary context is considered ($R^2$=0.442). Finally, we identify the specific behavioral and contextual variables that are driving predictions in order to gain preliminary insights into the underlying behavioral mechanisms.

\section{Methods}
\subsection{Sample and Data Collection}
We utilized archival behavioral user data from a sample of frequent Snapchat users collected through the Snapchat app as part of regular business operations. Frequent users were defined as users who used the app every day for the six-month period before data collection. Among the users who met this criterion, we drew a random sample (N=100,000; N=79,175 after cleaning; details described below; $M_{age}$ = 25.07;  $SD_{age}$ =  7.24; 56.31 \% female) and obtained their behaviors on the Snapchat app during the 30-day period from July 6th to August 5th, 2021. The dataset contained 105,636,289 sessions (1,334 sessions per user on average) which were distributed across 21,604,570 session hours (273 session hours per user on average). A session is defined as the interval between opening and closing the Snapchat app, and a session hour is defined as a full hour in which at least one session has occurred. 

All behavioral user data were obtained from in-app event logs indicating that a user had interacted with a specific feature of the app. The set of behavioral features included the number of sessions, time spent on the app, and users’ interactions with different features of the app (e.g., chats, Snaps, Stories, ads, Discover, Spotlight, creative tools, and Lenses), split by the type of interaction (view, send, create), and the source of the interaction (e.g., subscriptions, feed, reply; see SI 1). The total number of these behavioral features before further processing was 28 (see SI 2.1).

The set of context features included socio-demographic context, weather, temporal context, location visits, and connectivity status. This feature set was chosen to enable the models to learn a comprehensive representation of spatial context, including macro features that are relatively stable and largely beyond the users' control (e.g., socio-demographic features and weather at the ZIP code level), as well as more fine-grained features that reflect users' choices and mobility habits to a greater extent (e.g., location visits, connectivity status).
In order to enrich the behavioral data with socio-demographic and weather data, the GPS coordinates of each session were mapped to the corresponding ZIP code \cite{peters_big_2022}. Exact GPS coordinates were discarded in order to preserve users’ privacy. We then used the ZIP code to collect weather data and socio-demographic census data for the area a user was located in during a given session. Weather data was collected from OpenWeatherMap \cite{noauthor_openweathermap_2021}. We used OpenWeatherMap’s historical weather API to collect hourly weather data for ten raw features for each ZIP code: temperature, perceived temperature, atmospheric pressure in hPa, humidity in percent, minimum temperature, maximum temperature, wind speed in meter/sec, wind direction in degrees, cloudiness in percent, and categorical weather descriptors (e.g., sunny, rain, snow, fog, extreme, etc.). Socio-demographic data were obtained from the official website of the US Census \cite{noauthor_census_2021}. We captured 19 features belonging to several important socio-demographic categories, including socio-economic status, racial composition, age distribution, gender distribution, and marital status at the ZIP code level.
We also obtained location visitation data for users who agreed to share their location. To preserve users’ privacy, the locations were represented as 11 high-level categories, including events, travel, education, nightlife, residence, food/beverage, shops/services, arts/entertainment, outdoors/recreation, other, and missing. Location features were operationalized as the maximum probability score produced by the location classifier for each location category in each given session hour. The exact location names were not used, making it impossible to infer individual users’ exact location visits from the processed data. 
Additionally, we obtained connectivity status (Wi-Fi access, mobile data) as well as the temporal context (hour of the day, day of the week, day of the month, day of the year, time since the previous session) from the app usage logs (see SI 2.2).

The protocol was approved by the Columbia University IRB (AAAU2607). All methods were carried out in accordance with relevant guidelines and regulations. The study was exempt from the informed consent requirement by the Columbia University IRB because only anonymized archival data were used. While the research relies on proprietary data that cannot be shared openly, the code made available on this project's OSF page (\href{https://osf.io/nkfhz}{https://osf.io/nkfhz}) provides a detailed picture of the key properties of the dataset and data processing.

\subsection{Data Preprocessing and Operationalizations}
To facilitate training, hyperparameter tuning, and model evaluation, we split the data into three distinct datasets: a training set, a validation set, and a test set. The data was split at the person level, such that all records associated with any individual were assigned to only one of the three datasets. The training data consisted of 500,000 focal session hours, while the validation and test set each consisted of 50,000 focal session hours. Importantly, each focal session hour was associated with a history of up to 100 preceding session hours, each representing hundreds of sessions throughout the 30-day observation period.
The target variable, user engagement, was operationalized as the ratio of active-use scores and passive-use scores. Active-use scores were calculated as weighted averages of event counts indicating behaviors associated with active use. This includes the number of chat messages sent, the number of direct Snaps created or sent, and the number of Stories posted by a user. The different input scores were weighted using min-max-transformations such that they equally contributed to the active-use score, irrespective of their absolute frequencies. This approach ensures that relatively rare actions that require higher levels of effort and engagement (e.g., posting a story) would not be drowned out by more frequent but effortless actions (e.g., sending a message). Passive-use scores were defined as the number of Stories viewed. This includes Stories posted by friends, as well as curated Stories viewed through the Discover and Spotlight feeds. These definitions map onto the distinction between social active use and passive use as presented in the Passive and Active Facebook Use Measure (PAUM) \cite{gerson_passive_2017}.

Because user engagement was analyzed at the hourly level, we aggregated the raw behavioral event logs to hourly count metrics for all 28 relevant behavioral features. Since the behavioral count data tended to be power-law distributed, we transformed the hourly count data in several ways: First, we removed data points in the top 0.001 quantiles of each feature distribution in order to remove the most extreme outliers, which are likely caused by technical glitches or bots. Second, we normalized the feature values by dividing hourly counts by hourly active time on the app while also retaining the original non-normalized feature space. This was done to not only measure user activity in absolute terms but also how active users were relative to the time they spent on the app, which can be interpreted as a measure of intensity. Third, we log-transformed the combined feature set to produce more balanced distributions - again, while retaining the original non-log-transformed feature space. Additionally, we generated 15 derived features, such as composite scores of active behaviors (including chats, Snaps, and Stories created or shared by a user), composite scores of passive behaviors (Stories consumed through the Discover, Spotlight, and friends’ Story feeds), as well as several ratio and difference scores: the ratio and difference of chat messages sent and received, the ratio and difference of direct Snaps sent and received, the ratio and difference of Stories posted and viewed, as well as ratio and difference of active and passive use scores. Through this process, we obtained an overall space of 127 behavioral features (see SI 2.1).

Similar to behavioral features, context features were aggregated to the hourly level. For ZIP-code-level features (i.e., weather and socio-demographic census features), we calculated time-weighted averages (by active time on the app) if a user had visited more than one ZIP code within one hour. For location features, we cast each of the location categories as a feature and used the maximum probability score (per category, per hour) obtained from a location classifier. Missing values were imputed with zeros as the location classifier was not able to pick up on a relevant location in these cases. Connectivity status was derived as the ratio of the number of sessions involving mobile data usage and the number of sessions involving Wi-Fi access for each hour. It was not necessary to further aggregate the five temporal context features since none of them were captured at a more granular level than the hourly level. Overall, our model included 56 context features (see SI 2.2).

Finally, all numerical (interval and ratio scaled) features were normalized using min-max transformations, such that the final range for each feature was limited to values between 0 and 1 on the training set. Categorical (nominally and ordinally scaled) features were one-hot-encoded (dummy coded), increasing the number of weather features from 10 to 19. All transformations involving distribution parameters were fitted on the training set and then applied to the validation set and the test set using the distribution characteristics of the training set. The final feature set included 183 preprocessed predictors (see SI 2).

\subsection{Modeling}
In distinction from much of the behavioral science literature on user engagement and media and technology habits, we employ a predictive modeling approach \cite{hofman_prediction_2017, hofman_integrating_2021, yarkoni_choosing_2017, breiman_statistical_2001, shmueli_explain_2010}. The prediction of user engagement was framed as a multivariate time series regression problem, where each feature represented a temporal sequence of behaviors or context events, and the target variable was a continuous variable representing active and passive use. A neural network architecture that is particularly well suited to represent recurrent patterns in multivariate time series is the LSTM architecture \cite{hochreiter_long_1997, schmidhuber_deep_2015}. LSTM neural network models are a type of recurrent neural network (RNN), that is able to preserve information from previous steps more effectively by combining hidden states from previous time steps and current inputs before feeding them through a series of gates (input gate, forget gate, output gate), determining which information is passed on across time steps. In contrast to standard RNN models, which are limited to shorter time series, LSTM neural networks are able to represent the temporal order of widely separated events and temporally extended patterns in noisy input sequences and effectively extract information captured in the temporal distance between events in long sequences \cite{gers_learning_2000, hochreiter_long_1997}. In other words, LSTM models are designed to learn recurrent patterns over time and across long input sequences. LSTM models play an important role in a wide variety of applications, including time series prediction \cite{hua_deep_2019, siami-namini_comparison_2018}, anomaly detection \cite{malhotra_long_2015}, speech recognition \cite{graves_hybrid_2013}, and business process management \cite{camargo_learning_2019, tax_predictive_2017}. In the present project, we leverage the capacity of LSTM neural networks to make predictions from a large number of long time series. 

More concretely, we utilized a network architecture of stacked LSTM layers, where the output of each LSTM layer was fed into the next LSTM layer. The output of the last LSTM layer was fed into a final dense layer which was connected to a 1-neuron output layer with a linear activation function. All dense layers utilized rectified linear unit (relu) \cite{nair_rectified_2010} activation functions, and all LSTM layers utilized hyperbolic tangent (tanh) \cite{kalman_why_1992} activation functions. We used Bayesian optimization \cite{falkner_bohb_2018} to tune the depth of the neural network and the dimensionality of the layers. We also tuned the learning rate and the regularization strength with dropout and recurrent dropout. The hyperparameter search space spanned LSTM branch depths of 1-4 layers, tapered LSTM layer dimensions [32, 64, 128, 256] (depending on the depth of the network), top layer dimensions [32, 64, 128], dropout and recurrent dropout values [0, 0.1, 0.2, 0.3, 0.4, 0.5, 0.6], as well as learning rates  [0.01, 0.001, 0.0001]. The batch size was set to a value of 2048 across all experiments. A full overview of the hyperparameter space can be found in Table SI 3. To ensure that performance differences between the different model specifications were indeed caused by differences in the information content of the feature sets and not merely by a mismatch between the dimensionality of the feature space and network configuration, we re-tuned the hyperparameters for each learning task. Hyperparameter tuning was performed with Bayesian optimization \cite{falkner_bohb_2018} using three randomly selected starting points and 100 search iterations per model. Models were trained over 50 epochs with early stopping and a patience parameter value of 5. After training, we selected the model state resulting in the lowest Root Mean Squared Error (RMSE) on the validation set. Generalized model performance was assessed by re-fitting the model with the best hyperparameter configuration and evaluating its predictions on the test set. To ensure the robustness of these performance estimates, we repeated the procedure ten times for each model. An overview of the hyperparameter settings for each model can be found in SI 4.

\section{Results}
\subsection{Predicting Active and Passive Use from Histories of In-App User Behaviors (RQ1)}

To test whether active and passive use can be predicted from users’ histories of in-app behaviors, we trained an LSTM model using all 127 behavioral feature sequences as inputs (Model 1). We found an $R^2$ coefficient of 0.345, meaning that the model was able to explain 34.5\% of the variance in active-passive-use scores, which is considerably better than a naive baseline of $R^2$=0. The result indicates that patterns of past behavior are predictive of active and passive use at a given point in time, which is consistent with the idea of habit-driven patterns of engagement. A graphical representation of the results can be found in Figure \ref{fig:plot_rq1_r2} and detailed values can be found in SI 5 (Model 1).

\subsection{Effects of Context-Awareness on Model Performance (RQ2)}
\begin{figure}
  \centering
  \includegraphics[width=1\textwidth]{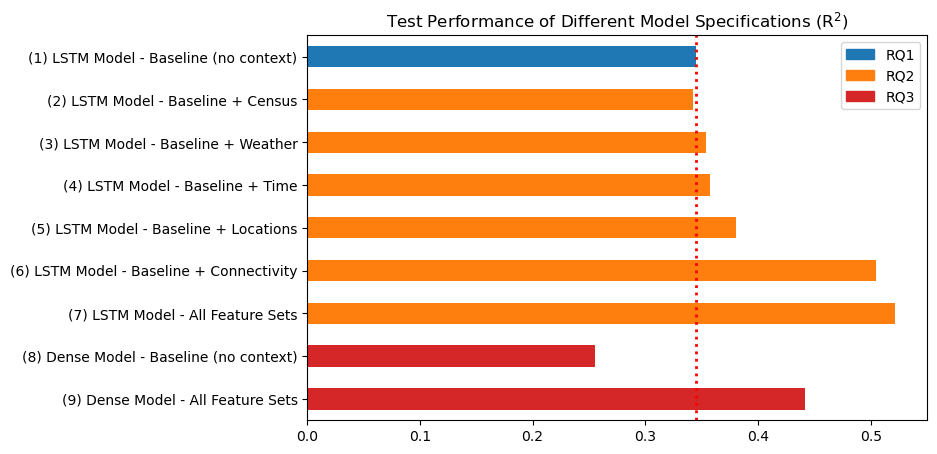}
  \caption{Overview of the predictive performance of different model specifications. RQ1: Model trained on behavioral histories only. RQ2: Model specifications used to assess the performance increment due to different sets of context features, including socio-demographic context (Model 2), weather (Model 3), temporal context (Model 4), location visits (Model 5), network connectivity status (Model 6), and all context features (Model 7). RQ3: Predictive performance of dense neural network models trained on cross-sectional data. Model 8 was trained on only behavioral data from t-1. Model 9 was trained on behavioral data from t-1 and context features from t0. For more details, please see SI 5.}
  \label{fig:plot_rq1_r2}
\end{figure}

In order to assess whether context-awareness can contribute to the prediction of user engagement, we employed LSTM neural networks in conjunction with ablation study techniques. In particular, we defined six additional model specifications that gave the model access to different subsets of context features and allowed us to observe how model performance changed in response to the addition/removal of context features. We used the previous model consisting of only behavioral sequences as a baseline. We then added each of the five individual brackets of context features (socio-demographic context, weather, temporal context, location visitations, connectivity status) to the baseline model in order to assess the increase in predictive performance associated with each of the different types of context features independently. Finally, we added all context features at once in order to assess the overall increase in predictive performance. An overview of the learning tasks, feature sets, and results can be found in Figure \ref{fig:plot_rq1_r2} and SI 5.

The results show that weather (Model 3, $R^2$=0.354) and temporal context (Model 4, $R^2$=0.357) were associated with small performance improvements (2.5\% and 3.5\%, respectively), while location information (Model 5, $R^2$ =0.380) was associated with a moderate performance improvement of 10\% over the baseline (Model 1). With a performance improvement of 46\%, connectivity status (Model 6, $R^2$ =0.504) added by far the most incremental explained variance. Adding census data to the baseline model did not result in improved performance (Model 2, $R^2$=0.342). Adding all features to the baseline simultaneously (Model 7, $R^2$=0.522) resulted in the best model and the largest performance improvement (51\%), indicating that the different feature sets capture some non-redundant information. The model also dramatically outperformed models trained exclusively on context features (see Section 3.3). These findings suggest that complementing behavioral features with context features, especially connectivity status, can substantially improve the performance of models predicting user engagement. The findings are consistent with the idea of context-contingent patterns of engagement.

% TODO remove references to RQs

\subsection{Effects of Context in Models Trained on Truncated Sequences (RQ3)}
\begin{figure}
  \centering
  \includegraphics[width=1\textwidth]{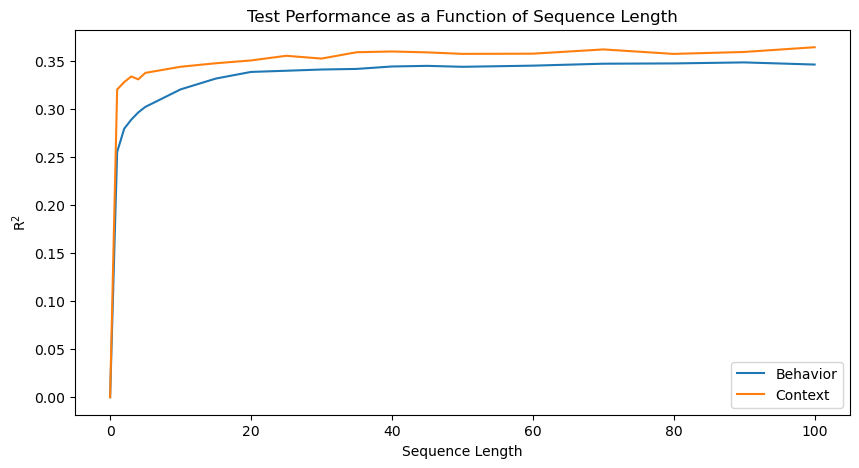}
  \caption{Model performance as a function of sequence length for behavioral histories (blue) and context histories (orange). The curves show diminishing returns to adding additional time steps to the model. Models built on behavioral histories benefit more from long time series compared to models built on context histories.}
  \label{fig:sequence_lengths}
\end{figure}

In order to assess the history-dependence of the predictive models, we conducted an ablation study in which we manipulated the length of the sequence that was accessible to the LSTM model. Sequence lengths ranged from only 1 to 100 time steps, representing anything from just one session up to 100 session hours distributed across 28 days. We first conducted this analysis for behavioral sequences and then separately for sequences of context data. In the case of behavioral histories, the longest sequence ranged from t-101 to t-1. No information about t0 was included in order to ensure that no information regarding user activity in the target session leaked into the predictor set. In the case of context histories, the longest sequence ranged from t-100 to t0, such that the last data point in the sequence represented the momentary context in t0. Model performances are depicted in Figure \ref{fig:sequence_lengths}.

As expected, longer time series were generally associated with higher model performance. This effect was more pronounced in the case of behavioral histories compared to context histories. For example, a model trained on behavioral histories of sequence length 1 captured 73\% of the maximum performance, while a model trained on context histories of sequence length 1 captured 88\% of the maximum performance. Models utilizing a sequence length of 50 yielded at least 99\% of the performance reached by models trained on the maximum length sequences with 100 time steps. The results indicate that temporal histories capture information about user engagement and that a relatively large share of variance in user engagement can be explained with very short sequences, particularly in the case of context predictors.

The previously trained models rely on sequences of historical data to predict user engagement. While the findings suggest that even short sequences of behavioral and contextual data offer moderate levels of predictive performance, we designed an additional experiment to test the incremental predictive power of context data at the extreme. Specifically, we trained a series of models based on minimal historical data, including only cross-sectional behavioral features from t-1 and momentary context features from t0. Given that this procedure eliminates the need for time series analyses, we shifted the analytical strategy from the LSTM architecture to a feed-forward neural network architecture. 

The results show a substantial drop in performance when only behavioral features from t-1 are provided to the model. With an $R^2$=0.256, the performance of the model incorporating cross-sectional behavioral data (Model 8) was 26\% lower compared to the baseline model incorporating behavioral histories of sequence length 50 and 51\% lower compared to the best model incorporating the full feature set at a sequence length of 50. The cross-sectional model trained on behavioral and context features (Model 9), on the other hand, showed a predictive performance of $R^2$=0.442. This was 15\% lower than that of the best model from the previous analysis but 28\% higher than the performance of the baseline model with full access to behavioral histories and no context features. Most importantly, predictive performance was 73\% higher than the performance of the model built on only cross-sectional behavioral data. The results demonstrate that the prediction of user engagement - even in the extreme case of fully truncated time series - works reasonably well if context is considered. For a visual comparison, see Figure \ref{fig:plot_rq1_r2}.

\subsection{Model Explainability Analyses (RQ4)}
While deep neural networks excel at making predictions from complex data, they do not provide much insight regarding the strength and directionality of specific feature-target associations by default. In order to open up the black box and explain how the models arrive at their predictions, we employed Shapley values, a popular model explainability technique \cite{lundberg_unified_2017}. Shapley values are additive feature importance measures that quantify the contribution of each feature with respect to each individual predicted value. Intuitively, this is achieved by calculating the average marginal contribution of each feature across all possible subsets of features. The marginal contribution of a feature is defined as the performance difference between a model including the focal feature and a model withholding the feature (e.g., by replacing feature values with unrelated values sampled from a background dataset). Because the effect of withholding a feature is contingent on other features, the marginal effects need to be computed and averaged for all possible subsets of features. This procedure is repeated for each feature in the overall feature set. In the present project, we used the Shapley Additive Explanations (SHAP) \cite{lundberg_unified_2017} Python library. Due to their computational complexity, SHAP values were generated for the dense neural network model incorporating the full feature set (Model 9) on a subsample of 20,000 data points from the test set.

\begin{figure}
  \centering
  \includegraphics[width=1\textwidth]{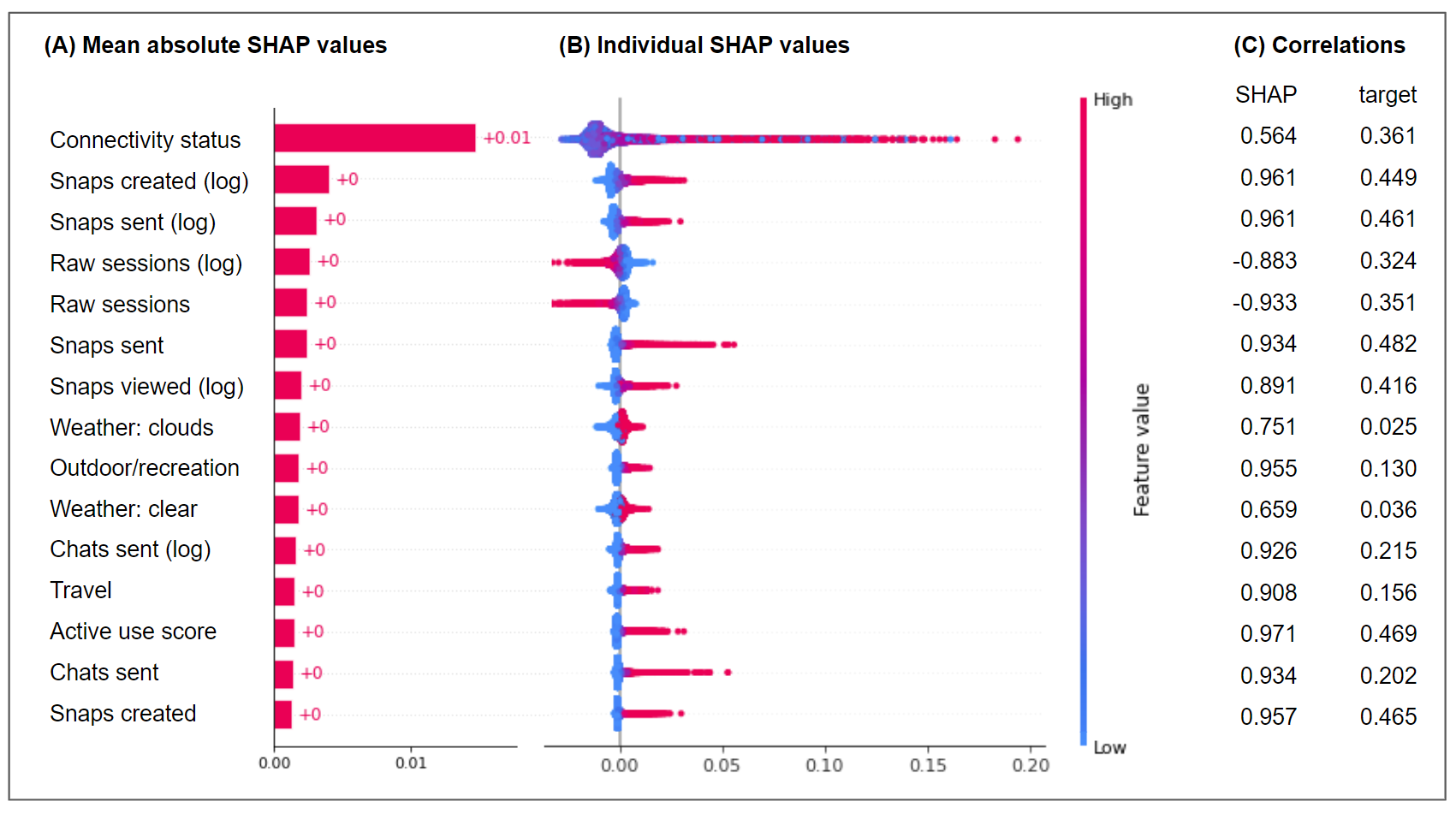}
  \caption{Mean absolute SHAP values (panel A), distribution of individual SHAP values colored by underlying feature values (panel B), and correlations between feature values and SHAP values, as well as feature values and target values (panel C). Connectivity status, online behaviors in the previous session, location visits, and weather features showed the largest impact on predictions.}
  \label{fig:shap}
\end{figure}

 In order to estimate general feature importance scores, we computed the arithmetic mean of the absolute Shapley values for each feature across all data points in the test sample. To analyze the directionality of the association, we analyzed the relationship between feature values and Shapley values. For comparison, we also analyzed the relationships between feature values and actual target values on the test set using Pearson’s correlation coefficients for metric features and point-biserial correlation coefficients for one-hot encoded categorical features. A positive correlation between feature values and Shapley values indicates that higher feature values are, on average, associated with higher predicted values in the model. A positive correlation between feature values and actual target values would indicate linear feature-target associations irrespective of their representations in the model. The results are shown in Figure \ref{fig:shap}.

Connectivity status and location visits in t0, along with in-app behaviors related to active use in t-1, were particularly important predictors. Connectivity status showed by far the greatest average SHAP value. Mobile data usage was overall associated with higher predictions of active use. A closer inspection of the SHAP values revealed that, in some cases, feature values for high mobile data usage were associated with negative SHAP values. This indicates that contrary to the general trend, mobile data usage sometimes decreases predictions of active use, meaning the model represents interaction effects or nonlinearities. The idea of interactive relationships is also reflected in another finding: despite its high overall impact on predicted values, the correlations between connectivity status and SHAP values, as well as actual target values, are lower than those of the behavioral predictors. This is likely the case because correlations, as opposed to SHAP values, only represent linear relationships. 

\section{Discussion}
\subsection{Interpretation of Results}
The results show that user engagement is highly predictable and that context-awareness can substantially improve the performance of models predicting patterns of active versus passive use compared to models trained on behavioral histories alone, especially when minimal temporal histories are considered. While the baseline model explained 34.5\% of the variation in active-passive use scores, the context-aware model, including all context features, explained 52.2\%, amounting to a performance improvement of 51\%. The performance improvement was even more pronounced in the models trained on truncated sequences, where the context-aware model scored 73\% higher than the behavioral baseline model. These findings underscore the value of contextualized representations of user behavior for predicting user engagement on social platforms. Additionally, our findings are aligned with previous research suggesting that user behavior on Snapchat follows habitual patterns over time \cite{chowdhury_ceam_2021, liu_characterizing_2019}. More broadly, our findings are consistent with the notion that social sharing \cite{choi_understanding_2021, bayer_texting_2012} and media consumption \cite{anderson_habits_2021, diddi_getting_2006, wood_habits_2002, bayer_connection_2016, peters_social_2024} are habit-driven. 

In line with the idea that specific behavioral outcomes are best predicted by constructs that match their level of analysis, we found that connectivity status (specific to the individual user’s device) and location (an individual user’s immediate surroundings) were associated with the greatest uptick in predictive performance, while less specific context factors operating at larger geographic scales - such as weather or socio-economic context - were of little help. The small effect of weather on predictive performance is consistent with previous work suggesting that weather has only small effects on psychological states and behavioral outcomes \cite{harley_psychology_2018}.

The finding that behavioral and contextual histories from the recent past were more predictive than information from the distant past is plausible, considering that the effects of predictors captured at a more recent point in time are more direct and less likely to be washed out by noise. Additionally, more recent predictor states contain information about previous states, such that adding previous states to a model will only marginally improve model performance insofar as they contain non-redundant information. At the same time, the finding reflects a general limitation of LSTM models, which place more emphasis on more recent data points due to information decay \cite{chien_slower_2021}. The fact that the effect was more pronounced in the case of behavioral histories, however, speaks to the relevance of longer-term behavioral patterns, contrasted with more temporally bounded context effects.

The results of the model explainability analysis using SHAP values are aligned with the findings of the previous ablation study, indicating that connectivity status and location features are particularly important context factors. Mobile data usage was generally associated with active use. This is plausible given that people might limit passive use in order to save mobile data, but more importantly, people likely encounter more situations that are worth sharing when they are out and about as compared to when they are at home or at work, where they would typically have Wi-Fi access. The latter point is especially plausible given the image-based nature of Snapchat. The finding also ties in with previous work suggesting that social sharing is habitual \cite{choi_understanding_2021}. This interpretation is aligned with the finding that locations related to outdoor/recreation and travel are also associated with higher levels of active use in the present study. The fact that high feature values were not uniformly associated with high SHAP values is consistent with the idea that the neural network model represents interactive relationships and that those interactive relationships are of particular importance for the prediction of user engagement.

Taken together, our research extends previous work in several ways: First, we add nuance to the study of user engagement by focusing on the distinction between active and passive use, which had not been studied through the lens of media and technology habits before \cite{escobar-viera_passive_2018, hemmings-jarrett_evaluation_2017, khan_social_2017, pagani_influence_2011, trifiro_social_2019}. Second, we use highly granular large-scale user data to explore the effects of a wide range of objectively captured context factors, contrasting their predictive contributions against each other and against behavioral features. In doing so, our study pushes the boundaries of both research on media and technology habits and research on context-aware modeling of user engagement. Finally, our study demonstrates how integrating objective sensing and location-tracking can enrich behavioral research. By taking research out of the lab and into the real world, we glean insights that advance behavioral theory while also speaking to people's naturalistic experiences in everyday life, thus enhancing the ecological validity and applicability of the findings.

\subsection{Limitations and Directions for Future Research}

Our work has several limitations that have the potential to stimulate future research. First, the present study squarely follows a predictive modeling approach. This approach contrasts with explanatory methods, typically used in the behavioral and social sciences, that employ statistical inference to uncover causal relationships between variables \cite{breiman_statistical_2001, shmueli_explain_2010, hofman_integrating_2021, hofman_prediction_2017, yarkoni_choosing_2017}. While our methodology allows us to investigate the predictive contributions across a range of theoretically informed feature sets based on highly granular longitudinal user data, it inherently prioritizes prediction at the expense of causal explanation. Recognizing this distinction, future research could build on our findings to derive more targeted research questions and focus on specific causal mechanisms.

Second, while we interpret the presence of recurrent predictable patterns of user engagement through the lens of media and technology habits, our findings also relate to other theoretical orientations such as ecological psychology \cite{barker_explorations_1965, heft_ecological_2016, lobo_history_2018} and person-situation interactions \cite{rauthmann_person-situation_2020, funder_persons_2009, fleeson_end_2008}. Although our research was not designed as an explicit test of theoretical predictions from either framework, our findings are aligned with these theoretical traditions. Specifically, they add to the existing literature by examining the predictive power of context in the setting of online social platforms and by focusing on the prediction of user engagement from highly granular behavioral data based on recurrent patterns. By offering empirical support for the existence of interactive relationships between user engagement and certain context features, our findings corroborate the main premise of person-situation interactions. However, additional research is needed to disentangle these relationships in a more explicit manner.

Third, while our study leverages an exceptionally granular and comprehensive feature set, we encountered several tradeoffs with respect to the operationalization of context. For instance, integrating network-related features to capture user relationships \cite{baten_predicting_2023} could have enriched our analysis by offering deeper insights into the predictive utility of social contexts. Additionally, the location visits, a critical component of our analysis, were inferred from noisy data. This limitation was compounded by our decision to use broad location categories in order to safeguard user privacy, potentially at the expense of achieving maximum predictive performance. Furthermore, past research has highlighted the value of incorporating self-reported measures to capture the psychological experience of context \cite{schoedel_snapshots_2023, rauthmann_conceptualizing_2021, rauthmann_situational_2014}, a methodological approach we could not employ due to our reliance on objective behavioral data. Future work should continue to integrate objectively measured cues with self-report measures to gain deeper insights into psychological mechanisms.

Fourth, we computed active-passive use scores as the ratio of active and passive behaviors within a given session hour. This decision was necessary given that active and passive use in isolation are highly correlated with overall app use, rendering the prediction task trivial and less interesting from a theoretical perspective. Our goal was to create a measure that captures users' relative tendency towards either of the usage modes. Future work could avoid this tradeoff by shifting the level of analysis to that of individual user behaviors.

Fifth, past research has sometimes treated preceding behaviors or activities as a specific type of context \cite{rauthmann_conceptualizing_2021, wood_new_2007}. We fundamentally agree with this perspective but decided to emphasize the dichotomy of spatial context and behavioral histories as the distinction maps more cleanly onto current approaches in user modeling, which has often relied exclusively on past behavior \cite{purificato_user_2024, webb_machine_2001}.

Finally, our research was conducted on a single online social platform with unique affordances \cite{bucher_affordances_2018} and in a specific cultural context \cite{alsaleh_cross-cultural_2019}. Additionally, we only included data from adult users, which excludes a substantial share of Snapchat's user base. As Snapchat shares important features, such as the specific modalities for active and passive use (Chat, Stories, Spotlight, Discover) with other online social platforms, we believe it is likely that our findings would apply to related settings. However, the extent of generalizability would have to be examined in future research.

\subsection{Implications}
Our findings indicate that active and passive use follow predictable patterns over time and that context-awareness can substantially improve model performance. This has implications for the use of predictive models and their integration into the design of online platforms. Being able to accurately predict user behavior through context-aware modeling could facilitate the dynamic allocation of computational resources to adaptively support different modes of user engagement. For example, knowing when an app user likes to consume rather than create content would enable a platform to allocate resources to a recommender system and preload relevant content. Similarly, knowing when a user is likely to actively create and share content would allow them to pre-allocate resources to the camera process and load the user’s favorite filters. At the same time, a better understanding of the relationships between context factors and user engagement could facilitate the design of features that encourage active or passive use by making specific context factors salient to the user. For example, a system that draws users’ attention to the social affordances of their environment might encourage active use and discourage passive use in certain settings. While the present paper focuses on active and passive use, psychologically informed context-aware models can be used to personalize a wide array of features related to users’ momentary states, including physiological (e.g., being hungry or tired), social (e.g., being with family, friends, or work colleagues), or psychological states (e.g., mood and affect). Such personalizations that take into account context-contingent states will ultimately improve user experience and create new opportunities for online platforms to generate revenue.

Importantly, our findings can help decision-makers navigate resource and privacy tradeoffs. The ablation study that was used to analyze performance as a function of the length of data histories shows strongly diminishing returns to longer time series, especially in excess of sequence lengths of 50 time steps. While the specific relationship is highly dependent on the predictive task and the model at hand, our procedure constitutes a simple method to assess a model's dependence on data histories that can easily be used to inform decisions in practice to preserve resources and limit privacy concerns. Additionally, our results show that the integration of context data can compensate for the truncation of behavioral histories, offering an opportunity to fade out models relying on historical data in favor of more ephemeral modeling approaches representing momentary user states. Accordingly, in situations where maximizing predictive accuracy is not an absolute imperative, it might be recommended to use short-term behavioral data in conjunction with easily obtainable context features, such as connectivity status, to build models that are privacy-preserving and cheap to run. Smaller, less computationally expensive models can, in turn, unlock additional privacy benefits - for example, it would be possible to run models locally on a mobile device without sharing sensitive data.

Taken together, our findings challenge the narrative that long-term user data is needed in order to make accurate predictions of user engagement and provide personalized services \cite{jannach_recommender_2016, najafabadi_survey_2019, webb_machine_2001}. Instead, our results show that it is possible to trade a relatively small reduction in predictive performance for a considerable reduction in data requirements and computational resources. Practitioners should be aware of this fact in order to make informed decisions and balance considerations of predictive performance, resources, and privacy when implementing predictive models in real-world settings.

\subsection{Conclusion}
In conclusion, our findings show that active and passive use follow predictable patterns over time and emphasize the importance of context in predicting user engagement on social platforms. Consistent with theories on media and technology habits, the combination of behavioral data with contextual information leads to a substantial increase in the predictive performance of user engagement. The present paper presents the first rigorous large-scale assessment of the predictive power of a wide array of context factors with respect to active and passive use on mobile social platforms. Our findings have potential implications for managerial, engineering, and design decisions - specifically with respect to the dynamic allocation of computational resources to adaptively support different modes of user engagement, privacy-preserving modeling of user behavior, and the design of features that encourage active or passive use. Furthermore, our work provides a starting point for a broader research program investigating additional context factors, different types of user engagement, and the underlying psychological mechanisms.

\newpage

\section*{Ethics approval and consent to participate}
The data collection was approved by the Columbia University IRB (AAAU2607). All methods were carried out in accordance with relevant guidelines and regulations. The study was exempt from the informed consent requirement by the Columbia University IRB because only anonymized archival data was used.

\section*{Consent for publication}
Not applicable.

\section*{Availability of data and materials}
The datasets generated and/or analyzed during the current study are not publicly available due to their highly sensitive, proprietary nature. Requests can be granted upon approval by Snap Inc.

\section*{Competing interests}
YL, FB, and MWB are employed by Snap Inc. The research was funded by Snap Inc. HP, RAB, and SCM declare no potential conflicts of interest.

\section*{Funding}
The research was funded by Snap Inc. 

\section*{Author contributions}
HP, YL, and MWB developed the research idea and the research design. HP and YL collected and analyzed the data. HP wrote the paper. YL, FB, RAB, SCM, and MWB provided feedback on research design, analyses, and writing.

\section*{Acknowledgements}
We thank Ron Dotsch and Joseph B. Bayer for their thoughtful feedback. 

\newpage
\printbibliography

\end{document}

\maketitle
\vspace{5}
% % keywords can be removed
% \keywords{Context-Aware Modeling \and User Engagement \and Active and Passive Use \and ML \and LSTM \and Snapchat}

% \newpage
\section{Description of the Key Features of the Snapchat App}
\label{app:snapchat}

Snapchat is a major instant messaging platform with 397 million daily active users. In the US, the majority (65\%) of 18-29-year-olds use Snapchat, most of them at least daily. The key features of the Snapchat app are: 

\begin{itemize}
    \item Chat: The Chat feature allows users to send private messages that disappear after they have been viewed. Chats can be between two individuals or group chats.
    \item Snaps: Snaps are the most important content unit on Snapchat. Snaps are pictures or short videos that users send to one or more of their friends. Snaps can be sent privately, or they can be posted publicly as Stories.
    \item Stories: Stories are one or multiple Snaps that are shared with all of a user's friends or subscribers. Stories can be watched repeatedly and are deleted after 24 hours.
    \item Discover: The Discover feed contains curated Stories from publishers and creators, irrespective of whether a user has subscribed to them.
    \item Spotlight: Users can publicly share video content that shows up on the Spotlight feed of other users. Spotlight videos are presented to users based on a personalized recommender system.
    \item Creative Tools: Snapchat allows users to apply Filters (image effects) or Lenses (AR effects) to their Snaps (photos and videos).
    \item Map: The Map feature allows users to see their friends’ activity mapped across geographic locations.
    \item Camera: The camera screen allows users to take Snaps and access Creative Tools.
\end{itemize}
\newpage

\FloatBarrier

\begin{table}[H]
\scriptsize
\onehalfspacing

\vspace{0.5cm}
\section{List of Features}
\label{app:features}
\subsection{Behavioral Features}
\vspace{0.5cm}
        \centering
    \begin{tabularx}{\textwidth}{p{2.5cm} p{5cm} X} 
    \toprule
    Feature Type & Feature Name & Description \\
    \midrule
    Behavior count & raw\_session\_cnt & Number of raw sessions in given session hour \\
    Behavior count & session\_time & Overall session time in given session hour \\
    Behavior count & app\_open\_cnt & Number of times user opened the app in given session hour \\
    Behavior count & app\_open\_from\_notify\_cnt & Number of times user opened the app from a notification in given session hour \\
    Behavior count & chat\_view\_cnt & Number of chats viewed in given session hour \\
    Behavior count & chat\_send\_cnt & Number of chats sent in given session hour \\
    Behavior count & chat\_create\_cnt & Number of chats created (not necessarily sent) in given session hour \\
    Behavior count & chat\_snap\_view\_cnt & Number of snaps viewed in chat in given session hour \\
    Behavior count & direct\_snap\_create\_cnt & Number of direct snaps created in given session hour \\
    Behavior count & direct\_snap\_send\_cnt & Number of direct snaps sent in given session hour \\
    Behavior count & direct\_snap\_view\_cnt & Number of direct snaps viewed in given session hour \\
    Behavior count & direct\_snap\_send\_feed\_cnt & Number of direct snaps sent from feed in given session hour \\
    Behavior count & direct\_snap\_send\_camera\_cnt & Number of direct snaps sent from the camera in given session hour \\
    Behavior count & direct\_snap\_send\_reply\_cnt & Number of direct snaps sent as a reply to a message in given session hour \\

    % \end{tabularx}
    % \end{table}

    % \begin{table}[H]
    % \centering
    % \begin{tabularx}{\textwidth}{p{2.5cm} p{5cm} X} 
    
    Behavior count & direct\_snap\_send\_chat\_cnt & Number of direct snaps sent in chat in given session hour \\
    Behavior count & story\_snap\_post\_cnt & Number of story snaps posted in given session hour \\
    Behavior count & story\_snap\_view\_cnt & Number of story snaps viewed in given session hour \\
    Behavior count & story\_story\_view\_cnt & Number of stories viewed in given session hour \\
    Behavior count & story\_snap\_feed\_view\_cnt & Number of story snaps viewed in the feed in given session hour \\
    Behavior count & discover\_snap\_feed\_view\_cnt & Number of discover snaps viewed in feed in given session hour \\
    Behavior count & discover\_snap\_for\_you\_view\_cnt & Number of recommended discover snaps viewed in given session hour \\
    Behavior count & discover\_snap\_subscription\_view\_cnt & Number of discover snaps viewed from subscribed users in given session hour \\
    Behavior count & discover\_snap\_friends\_view\_cnt & Number of discover snaps viewed from friends in given session hour \\
    Behavior count & discover\_snap\_view\_cnt & Number of discover snaps viewed in given session hour \\
    Behavior count & spotlight\_view\_cnt & Number of spotlight videos viewed in given session hour \\
    Behavior count & creative\_tools\_open\_cnt & Number of times the creative tools page was opened in given session hour \\
    Behavior count & creative\_tools\_pick\_cnt & Number of times the user picked a creative tool in given session hour \\
    Behavior count & filter\_lens\_swipe\_cnt & Number of lenses swiped through in given session hour \\
    Behavior count & filter\_filter\_swipe\_cnt & Number of filters swiped through in given session hour \\
    Composite score & chat\_act\_pass\_ratio & Ratio of chats sent vs chats received in given session hour \\
    Composite score & chat\_act\_pass\_diff & Difference of chats sent vs chats received in given session hour \\
    Composite score & snap\_act\_pass\_ratio & Ratio of snaps sent vs snaps received in given session hour \\
    Composite score & snap\_act\_pass\_diff & Difference of snaps sent vs snaps received in given session hour \\
    Composite score & story\_act\_pass\_ratio & Ratio of stories posted vs stories viewed in given session hour \\
    Composite score & story\_act\_pass\_diff & Difference of stories posted vs stories viewed in given session hour \\
    Composite score & comp\_score\_act & Composite score of active behaviors in given session hour \\
    Composite score & comp\_score\_pass & Composite score of passive behaviors in given session hour \\
    Composite score & comp\_score\_create & Composite score of creative behaviors in given session hour \\
    Composite score & act\_pass\_ratio & Ratio of active composite score vs passive composite score in given session hour \\
    Composite score & act\_pass\_diff & Difference of active composite score vs passive composite score in given session hour \\
    Composite score & create\_pass\_ratio & Ratio of creative composite score vs passive composite score in given session hour \\
    Composite score & create\_pass\_diff & Difference of creative composite score vs passive composite score in given session hour \\
    Composite score & act\_create\_pass\_ratio & Ratio of active and creative composite score vs passive composite score in given session hour \\
    Composite score & act\_create\_pass\_diff & Difference of active and creative composite score vs passive composite score in given session hour \\
    \bottomrule
    \end{tabularx}
\end{table}

\FloatBarrier

\begin{table}[H]
\subsection{Context Features}
\scriptsize
\onehalfspacing
\vspace{0.5cm}
    \centering
    \begin{tabularx}{\textwidth}{p{3cm} p{4cm} X}
        \toprule
        Feature Type & Feature Name & Description \\
        \midrule
        Weather & temp & Temperature in Kelvin \\
        Weather & feels\_like & Temperature in Kelvin accounting for the human perception of weather \\
        Weather & pressure & Atmospheric pressure in hPa \\
        Weather & humidity & Humidity in \% \\
        Weather & temp\_min & Minimum temperature within the area in Kelvin \\
        Weather & temp\_max & Maximum temperature within the area in Kelvin \\
        Weather & wind\_speed & Wind speed in meter/sec \\
        Weather & wind\_deg & Wind direction, degrees (meteorological) \\
        Weather & clouds & Cloudiness in \% \\
        Weather & weather\_label\_clear & Weather labeled as clear \\
        Weather & weather\_label\_haze & Weather labeled as haze \\
        Weather & weather\_label\_rain & Weather labeled as rain \\
        Weather & weather\_label\_mist & Weather labeled as mist \\
        Weather & weather\_label\_smoke & Weather labeled as smoke \\
        Weather & weather\_label\_snow & Weather labeled as snow \\
        Weather & weather\_label\_clouds & Weather labeled as clouds \\
        Weather & weather\_label\_fog & Weather labeled as fog \\
        Weather & weather\_label\_drizzle & Weather labeled as drizzle \\
        Weather & weather\_label\_dust & Weather labeled as dust \\
        Socio-demographic & people\_per\_unit & Average inhabitants per housing unit in a given ZIP code \\
        Socio-demographic & male\_perc & Percentage of male inhabitants in a given ZIP code \\
        Socio-demographic & race\_white\_perc & Percentage of residents identifying as white in a given ZIP code \\
        Socio-demographic & race\_black\_perc & Percentage of residents identifying as black in a given ZIP code \\
        Socio-demographic & race\_native\_perc & Percentage of residents identifying as native in a given ZIP code \\
        Socio-demographic & race\_asian\_perc & Percentage of residents identifying as Asian in a given ZIP code \\
        Socio-demographic & race\_hispanic\_perc & Percentage of residents identifying as Hispanic in a given ZIP code \\
        Socio-demographic & race\_more\_perc & Percentage of residents identifying as more than one race in a given ZIP code \\
        Socio-demographic & age\_5\_to\_9\_perc & Percentage of residents aged 5 to 9 in a given ZIP code \\
        Socio-demographic & age\_10\_to\_14\_perc & Percentage of residents aged 10 to 14 in a given ZIP code \\
        Socio-demographic & age\_15\_to\_19\_perc & Percentage of residents aged 15 to 19 in a given ZIP code \\
        Socio-demographic & age\_20\_to\_24\_perc & Percentage of residents aged 20 to 24 in a given ZIP code \\
        Socio-demographic & age\_25\_to\_34\_perc & Percentage of residents aged 25 to 34 in a given ZIP code \\
        Socio-demographic & age\_35\_to\_44\_perc & Percentage of residents aged 35 to 44 in a given ZIP code \\
        Socio-demographic & age\_45\_to\_54\_perc & Percentage of residents aged 45 to 54 in a given ZIP code \\
        Socio-demographic & avg\_household\_size & Average household size in a given ZIP code \\
        Socio-demographic & med\_inc & Median income in a given ZIP code \\

        % \end{tabularx}
        % \end{table}

        % \begin{table}[H]
        % \centering
        % \begin{tabularx}{\textwidth}{p{2.5cm} p{5cm} X}
        
        Socio-demographic & marriage\_married & Percentage of residents who are married in a given ZIP code \\
        Socio-demographic & marriage\_never\_married & Percentage of residents who never married in a given ZIP code \\
        Temporal context & time\_delta & Time passed since the previous session hour \\
        Temporal context & session\_hourofday & Hour of the day \\
        Temporal context & session\_weekday\_num & Day of the week (Monday=1, Sunday=7) \\
        Temporal context & session\_dayofmonth & Day of the month \\
        Temporal context & session\_dayofyear & Day of the year \\
        Location & loc\_event\_prob & Maximum probability score of event visit during session hour (e.g., sports event, concert) \\
        Location & loc\_travel\_prob & Maximum probability score of visiting a travel-related location during session hour (e.g., train station, bus terminal) \\
        Location & loc\_education\_prob & Maximum probability score of visiting an education-related location during session hour (e.g., school, university) \\
        Location & loc\_nightlife\_prob & Maximum probability score of visiting a nightlife-related location during session hour (e.g., bar, nightclub) \\
        Location & loc\_residence\_prob & Maximum probability score of visiting a residential location during session hour \\
        Location & loc\_food\_beverage\_prob & Maximum probability score of visiting a location related to the food and beverage industry during session hour (e.g., restaurant) \\
        Location & loc\_shops\_services\_prob & Maximum probability score of visiting a shop or service location during session hour \\
        Location & loc\_arts\_entertainment\_prob & Maximum probability score of visiting a location related to arts or entertainment during session hour (e.g., gallery, cinema) \\
        Location & loc\_outdoors\_recreation\_prob & Maximum probability score of visiting a location related to outdoors and recreation during session hour (e.g., parks, trails) \\
        Location & loc\_other\_prob & Catch-all category for locations other than the ones listed previously \\
        Location & missing & Flag indicating missing location data \\
        \bottomrule
    \end{tabularx}
\end{table}

\FloatBarrier

\begin{table}[h]
\vspace{0.5cm}
\section{Overview of the Hyperparameter Search Space}
\label{app:search_space}
\vspace{0.5cm}
    \centering
    \begin{tabular*}{\textwidth}{@{\extracolsep{\fill}} lll}
        \toprule
        \textbf{Hyperparameter Name} & \textbf{Search Space} & \\
        \midrule
        LSTM layers depth & [1, 2, 3, 4] & \\
        Dense layers depth & [1, 2, 3, 4] & \\
        LSTM layer dimensions & [32, 64, 128, 256] & \\
        Dense layer dimensions & [32, 64, 128, 256] & \\
        LSTM branch dropout & [0, 0.1, 0.2, 0.3, 0.4, 0.5, 0.6] & \\
        Recurrent dropout & [0, 0.1, 0.2, 0.3, 0.4, 0.5, 0.6] & \\
        Dense layer dropout & [0, 0.1, 0.2, 0.3, 0.4, 0.5, 0.6] & \\
        Learning rate & [0.01, 0.001, 0.0001] & \\
        Batch size & 2048 & \\
        Max epochs & 50 & \\
        Patience & 5 & \\
        \bottomrule
    \end{tabular*}
\end{table}

\FloatBarrier

\section{Chosen Hyperparameters by Model}
\label{app:hyperparams}
\subsection{Hyperparameters of Model 1 (LSTM, Baseline)}

\begin{table}[h]

    \centering
    % \vspace{0.5cm}
    % \vspace{0.5cm}
    \begin{tabular*}{\textwidth}{@{\extracolsep{\fill}} lll}
        \toprule
        \textbf{Hyperparameter Name} & \textbf{Hyperparameter Value} & \\
        \midrule
        LSTM layers depth & 1 \\
        LSTM layer dimensions & 32 \\
        Top layer dimensions & 32 \\
        LSTM branch dropout & 0 \\
        Recurrent dropout & 0.6 \\
        Learning rate & 0.001 \\
        Batch size & 2048 \\
        Max epochs & 50 \\
        Patience & 5 \\
        \bottomrule
    \end{tabular*}
\end{table}

\FloatBarrier

\begin{table}[h]
    \centering
    \vspace{0.5cm}
    \subsection{Hyperparameters of Model 2 (LSTM, Baseline + Census)}
    \vspace{0.5cm}
    \begin{tabular*}{\textwidth}{@{\extracolsep{\fill}} lll}
        \toprule
        \textbf{Hyperparameter Name} & \textbf{Hyperparameter Value} & \\
        \midrule
        LSTM layers depth & 2 \\        
        LSTM layer dimensions & 64, 32 \\        
        Top layer dimensions & 32 \\        
        LSTM branch dropout & 0 \\        
        Recurrent dropout & 0.6 \\        
        Learning rate & 0.0001 \\        
        Batch size & 2048 \\        
        Max epochs & 50 \\        
        Patience & 5 \\
        \bottomrule
    \end{tabular*}
\end{table}

\FloatBarrier

\begin{table}[h]
    \centering
    \vspace{0.5cm}
\subsection{Hyperparameters of Model 3 (LSTM, Baseline + Weather)}
\vspace{0.5cm}

    \begin{tabular*}{\textwidth}{@{\extracolsep{\fill}} lll}
        \toprule
        \textbf{Hyperparameter Name} & \textbf{Hyperparameter Value} & \\
        \midrule
        LSTM layers depth & 4 \\        
        LSTM layer dimensions & 256, 128, 64, 32 \\        
        Top layer dimensions & 64 \\        
        LSTM branch dropout & 0 \\        
        Recurrent dropout & 0.6 \\        
        Learning rate & 0.001 \\        
        Batch size & 2048 \\        
        Max epochs & 50 \\        
        Patience & 5 \\
        \bottomrule
    \end{tabular*}
\end{table}

\FloatBarrier

\begin{table}[h]
    \centering
    \vspace{0.5cm}
\subsection{Hyperparameters of Model 4 (LSTM, Baseline + Temporal Context)}
\vspace{0.5cm}

    \begin{tabular*}{\textwidth}{@{\extracolsep{\fill}} lll}
        \toprule
        \textbf{Hyperparameter Name} & \textbf{Hyperparameter Value} & \\
        \midrule
        LSTM layers depth & 3 \\        
        LSTM layer dimensions & 128, 64, 32 \\        
        Top layer dimensions & 32 \\        
        LSTM branch dropout & 0 \\        
        Recurrent dropout & 0.6 \\        
        Learning rate & 0.01 \\        
        Batch size & 2048 \\        
        Max epochs & 50 \\        
        Patience & 5 \\
        \bottomrule
    \end{tabular*}
\end{table}

\FloatBarrier

\begin{table}[h]
    \centering
    \vspace{0.5cm}
\subsection{Hyperparameters of Model 5 (Baseline + Location)}
\vspace{0.5cm}

    \begin{tabular*}{\textwidth}{@{\extracolsep{\fill}} lll}
        \toprule
        \textbf{Hyperparameter Name} & \textbf{Hyperparameter Value} & \\
        \midrule
        LSTM layers depth & 2 \\        
        LSTM layer dimensions & 64, 32 \\        
        Top layer dimensions & 32 \\        
        LSTM branch dropout & 0.2 \\        
        Recurrent dropout & 0.4 \\        
        Learning rate & 0.001 \\        
        Batch size & 2048 \\        
        Max epochs & 50 \\        
        Patience & 5 \\
        \bottomrule

    \end{tabular*}
\end{table}

\FloatBarrier

\begin{table}[h]
    \centering
    \vspace{0.5cm}
\subsection{Hyperparameters of Model 6 (LSTM, Baseline + Connectivity Status)}
\vspace{0.5cm}

    \begin{tabular*}{\textwidth}{@{\extracolsep{\fill}} lll}
        \toprule
        \textbf{Hyperparameter Name} & \textbf{Hyperparameter Value} & \\
        \midrule
        LSTM layers depth & 4 \\        
        LSTM layer dimensions & 256, 128, 64, 32 \\        
        Top layer dimensions & 32 \\        
        LSTM branch dropout & 0 \\        
        Recurrent dropout & 0.6 \\        
        Learning rate & 0.01 \\        
        Batch size & 2048 \\        
        Max epochs & 50 \\        
        Patience & 5 \\
        \bottomrule
    \end{tabular*}
\end{table}

\FloatBarrier

\begin{table}[h]
    \centering
    \vspace{0.5cm}

    \subsection{Hyperparameters of Model 7 (LSTM, All Features)}
    \vspace{0.5cm}

    \begin{tabular*}{\textwidth}{@{\extracolsep{\fill}} lll}
        \toprule
        \textbf{Hyperparameter Name} & \textbf{Hyperparameter Value} & \\
        \midrule
        LSTM layers depth & 3 \\
        LSTM layer dimensions & 128, 64, 32 \\        
        Top layer dimensions & 32 \\        
        LSTM branch dropout & 0 \\        
        Recurrent dropout & 0.6 \\        
        Learning rate & 0.001 \\        
        Batch size & 2048 \\        
        Max epochs & 50 \\        
        Patience & 5 \\
        \bottomrule
    \end{tabular*}
\end{table}

\FloatBarrier

\begin{table}[h]
    \centering
    \vspace{0.5cm}
    \subsection{Hyperparameters of Model 8 (Dense, Baseline)}
    \vspace{0.5cm}

    \begin{tabular*}{\textwidth}{@{\extracolsep{\fill}} lll}
        \toprule
        \textbf{Hyperparameter Name} & \textbf{Hyperparameter Value} & \\
        \midrule
        Dense layers depth & 1 \\        
        Dense layer dimensions & 32 \\        
        Top layer dimensions & 64 \\        
        Dense layer dropout & 0 \\        
        Learning rate & 0.001 \\        
        Batch size & 2048 \\        
        Max epochs & 50 \\        
        Patience & 5 \\
        \bottomrule
    \end{tabular*}
\end{table}

\FloatBarrier

\begin{table}[h]
    \centering
    \vspace{0.5cm}
    \subsection{Hyperparameters of Model 9 (Dense, All Features)}
    \vspace{0.5cm}

    \begin{tabular*}{\textwidth}{@{\extracolsep{\fill}} lll}
        \toprule
        \textbf{Hyperparameter Name} & \textbf{Hyperparameter Value} & \\
        \midrule
        Dense layers depth & 2 \\        
        Dense layer dimensions & 64, 32 \\        
        Top layer dimensions & 64 \\        
        Dense layer dropout & 0 \\        
        Learning rate & 0.001 \\        
        Batch size & 2048 \\        
        Max epochs & 50 \\        
        Patience & 5 \\
        \bottomrule
    \end{tabular*}
\end{table}

\begin{table}[h]
\section{Detailed Results}
\label{app:results}

\vspace{5}

\FloatBarrier
\newpage
\centering
\scriptsize
\begin{tabularx}{\textwidth}{@{}*{10}{X}@{}}
\toprule
Model \# & Model \newline Architecture & Model \newline Specification & Behavior & Census & Weather & Time & Location & Connectivity & R2 \\
\midrule
1 & LSTM & Baseline \newline (no context) & X & & & & & & 0.345 (0.0006) \\
\addlinespace 
2 & LSTM & Baseline + \newline Census & X & X & & & & & 0.342 (0.003) \\
\addlinespace 
3 & LSTM & Baseline + \newline Weather & X & & X & & & & 0.354 (0.0022) \\
\addlinespace 
4 & LSTM & Baseline + \newline Time & X & & & X & & & 0.357 (0.0073) \\
\addlinespace 
5 & LSTM & Baseline + \newline Locations & X & & & & X & & 0.380 (0.0018) \\
\addlinespace 
6 & LSTM & Baseline +  \newline Connectivity & X & & & & & X & 0.504 (0.0017) \\
\addlinespace 
7 & LSTM & All features & X & X & X & X & X & X & 0.522 (0.0038) \\
\addlinespace 
8 & Dense & Baseline & X & & & & & & 0.256 (0.005) \\
\addlinespace 
9 & Dense & All Features & X & X & X & X & X & X & 0.442 (0.0012) \\
\bottomrule
\end{tabularx}
\vspace{0.5cm}
\caption*{Note: Overview of the predictive performance of different model specifications. RQ1: Model trained on behavioral histories only. RQ2: Model specifications used to assess the performance increment due to different sets of context features, including socio-demographic context (Model 2), weather (Model 3), temporal context (Model 4), location visits (Model 5), network connectivity status (Model 6), and all context features (Model 7). RQ 4: Predictive performance of dense neural network models trained on cross-sectional data. Model 8 was trained on only behavioral data from t-1. Model 9 was trained on behavioral data from t-1 and context features from t0.}
\end{table}